\PassOptionsToPackage{table}{xcolor}
\documentclass[sigconf,preprint]{acmart}
\AtBeginDocument{%
  }

\usepackage{longtable}
\usepackage{tabularx}
\usepackage{multirow}
\usepackage{booktabs}
\usepackage{adjustbox}
\usepackage{listings}
\usepackage{graphicx}
\usepackage{tcolorbox}
\tcbuselibrary{breakable,listings,skins}
\usepackage{makecell}
\usepackage{subcaption}
\usepackage{amsmath}
\usepackage{pifont}
\usepackage{longtable}
\usepackage{tabularx}
\usepackage{multirow}
\usepackage{booktabs}
\usepackage{adjustbox}
\usepackage{graphicx}
\usepackage{tcolorbox}
\usepackage{makecell}
\usepackage{subcaption}
\usepackage{amsmath}
\usepackage{pifont}
\usepackage{tcolorbox}
\tcbuselibrary{breakable}
\newcommand{\ours}{\textsc{FINDERS}}
\newcommand{\bqa}{\textsc{BehaviorQA}}

\begin{document}

\title{Rethinking User Encoders: Routing Diffusion States for Scenario-Adaptive User Representation}

\author{
Junru Zhang\textsuperscript{1,2,\textasteriskcentered},
Jinyong Wen\textsuperscript{2},
Bin Dou\textsuperscript{2},
Wuliang Huang\textsuperscript{2},
Baokun Wang\textsuperscript{2,\textdagger},
Yu Cheng\textsuperscript{2},
Weiqiang Wang\textsuperscript{2},
Cheng Peng\textsuperscript{1,\textdagger}
}

\thanks{$\ast$ This work was completed during Junru's internship at Ant Group.}
\thanks{$\dagger$ Corresponding authors.}

\affiliation{%
  \institution{
    \textsuperscript{1}Zhejiang University,
    \textsuperscript{2}Ant Group
  }
  \country{China}
}

\renewcommand{\shortauthors}{Zhang et al.}

\begin{abstract}
Existing large language model (LLM)-based user encoders serialize heterogeneous behavioral logs and compress them into static representations through autoregressive (AR) prediction. However, their unidirectional constraint limits bidirectional information flow, hindering cross-modal interactions and the joint preservation of multi-source signals. Consequently, the learned static representations struggle to accommodate the distinct evidence requirements of diverse downstream scenarios. We address these limitations by unlocking the iterative denoising process of diffusion language models (dLLMs) as a multi-granularity reconstruction space. Based on this insight, we propose \ours{} 
(\textbf{F}actor\textbf{I}zatio\textbf{N} of
\textbf{D}iffusion stat\textbf{E}s for
\textbf{R}outing across \textbf{S}cenarios), a router-equipped dLLM framework comprising four operations: \emph{denoise}, \emph{factorize}, \emph{route}, and \emph{compress}. Learnable meta-slots organize complementary denoising states into transferable cross-modal semantics and modality-specific behavioral factors. To provide grounded supervision, we construct \bqa{}, a large-scale corpus containing explicit behavioral rationales. Guided by scenario queries, a lightweight router dynamically emphasizes relevant slot directions via an input-level residual pathway before cross-source fusion and final compression. Extensive experiments and online A/B tests across ten industrial scenarios demonstrate that \ours{} consistently outperforms state-of-the-art methods in both unified and scenario-adaptive settings. This work effectively pushes the boundaries of dLLMs beyond language generation, demonstrating their immense potential as powerful user representation learners.
\end{abstract}



\maketitle

\section{Introduction}
\label{sec:intro}

User representation learning supports personalized decision-making across industrial applications, including recommendation~\cite{sun2019bert4rec,ren2024representation}, digital marketing~\cite{yuan2026query,zhou2019deep}, and risk management~\cite{li2026panther,yuan2026query,ni2018perceive,yang2023empowering}.
In large-scale payment platforms, heterogeneous user signals, such as transaction records, browsing histories, app usage, and structured attributes, are commonly compressed into compact embeddings that can be centrally stored and reused across downstream scenarios~\cite{ni2018perceive,hou2022towards,yang2023empowering}.
However,  the same continuously updated user profile may support substantially different decisions. For example, fraud detection emphasizes transaction irregularities~\cite{zhou2019deep,xie2022learning}, marketing decisions focus on recent purchases and campaign interactions~\cite{li2019multi,dou2025transferable,zhou2019deep}, while credit assessment relies more on stable financial evidence~\cite{yuan2026query}.
Therefore, a practical user representation must preserve behavioral evidence at multiple dimensions while allowing each scenario to selectively access the information relevant to its objective.

Recent large language model (LLM)-based user encoders~\cite{zhu2024liber,gao2025instruction,gong2025behavegpt,li2026panther} serialize heterogeneous behavior logs and have demonstrated effectiveness in capturing complex behavioral semantics. Most of these methods~\cite{zhu2024liber,gao2025instruction,he2025llm2rec,li2026panther} learn unified user representations through autoregressive (AR) next-token or next-behavior prediction. Despite this progress, their strictly causal, left-to-right attention restricts bidirectional information flow, preventing earlier tokens from attending to subsequent contexts~\cite{vaswani2017attention,behnamghader2024llm2vec,yuan2026decoder}. This unidirectional constraint limits the interaction and joint preservation of multi-source signals distributed across the sequence. As a result, critical cross-modal interactions and modality-specific information can be insufficiently preserved or even lost. Consequently, the learned representation struggles to accommodate the distinct evidence requirements of diverse downstream scenarios.

Diffusion large language models (dLLMs) provide a promising alternative to this unidirectional paradigm~\cite{austin2021structured,sahoo2024simple,nie2026large,ye2025dream,zhang2025diffusion}. Through iterative bidirectional denoising, dLLMs allow behavioral tokens to exchange information in both directions, making them naturally suited to heterogeneous and partially ordered user logs~\cite{sahoo2024simple,nie2026large,ye2025dream,zhang2025diffusion}. However, while bidirectional interaction resolves the issue of unidirectional information flow, it does not inherently guarantee a scenario-adaptive representation. A key question therefore remains: \emph{how can the dynamic denoising states of dLLMs be organized and selectively accessed to satisfy different downstream scenarios?}

To address this question, we rethink diffusion-based user representation learning from a \emph{reconstruction-space} perspective. We view different denoising states not merely as intermediate steps, but as different reconstruction views that capture distinct semantic directions. For instance, for a clean transaction behavior, a higher-noise state can reflect the abstract intent of "making a payment," whereas a lower-noise state reconstructs the concrete action of "opening the Alipay app to pay." Consequently, higher-noise states increasingly highlight macro-behavioral intents, whereas lower-noise states retain more fine-grained details. These diverse states form a multi-granularity semantic space. Within this space, we can enable different scenarios to emphasize the diverse semantic directions most relevant to their objectives, such as detailed transaction irregularities for fraud detection or macro-behavioral patterns for marketing decisions.

\begin{figure}
    \centering
    \includegraphics[width=1\linewidth]{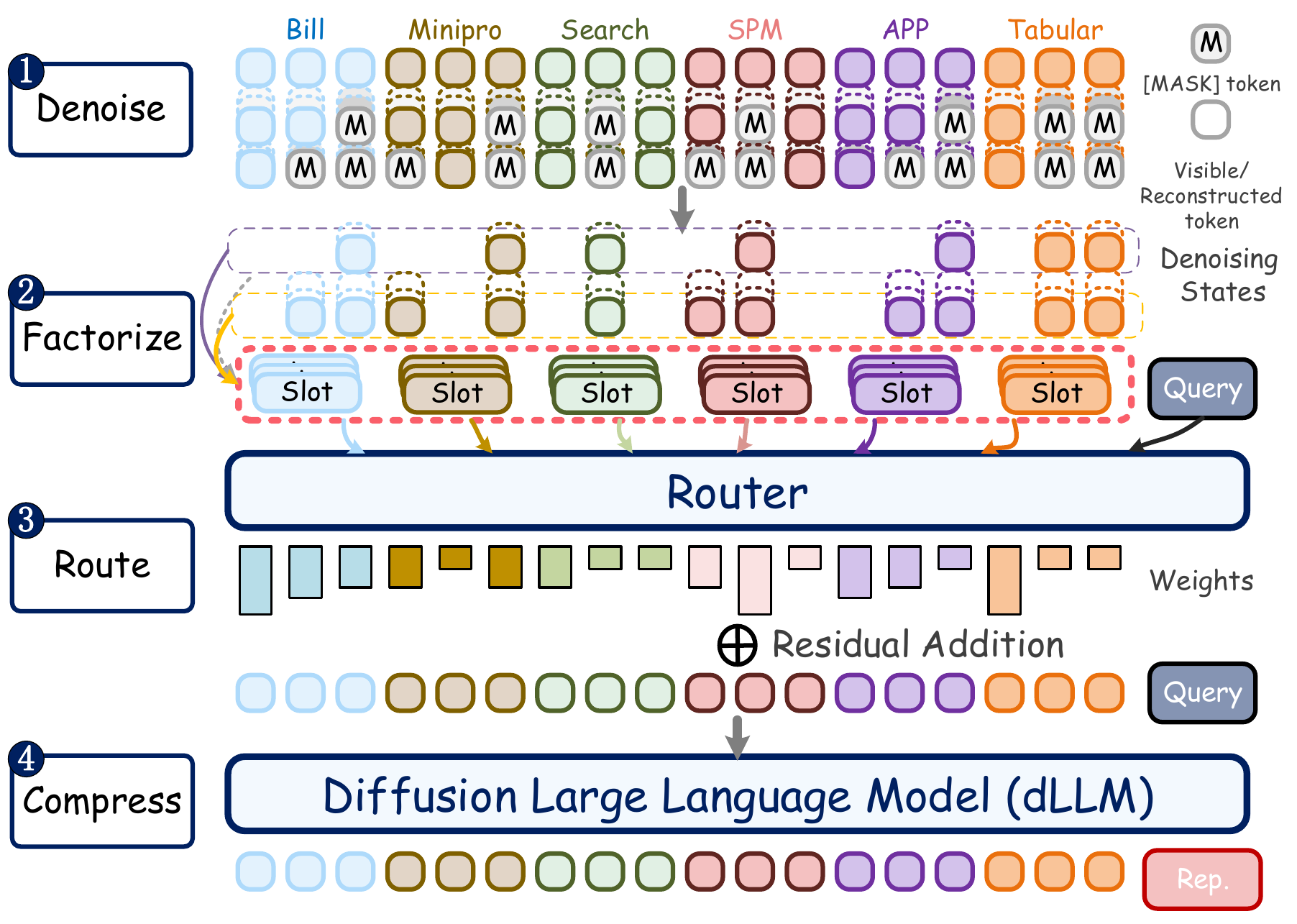}
\caption{\textbf{Overview of \ours{}.}
\ours{} first denoises heterogeneous behaviors into multi-granularity states and factorizes them into cross-modal and modality-specific slots. It then routes query-relevant slots to construct a residual enhancement and compresses the enhanced behavioral sequence into a scenario-adaptive representation.}
    \label{fig:arch}
    \vspace{-0.5cm}
\end{figure}

Based on this insight, we propose \textbf{\ours{}}
(\textbf{F}actor\textbf{I}zatio\textbf{N} of
\textbf{D}iffusion stat\textbf{E}s for
\textbf{R}outing across \textbf{S}cenarios), a router-equipped dLLM framework built around four synergistic operations: \emph{denoise}, \emph{factorize}, \emph{route}, and \emph{compress}. Fig.~\ref{fig:arch} illustrates the overall architecture of the proposed framework. Rather than directly pooling the final hidden states into a static embedding, \ours{} employs learnable meta-slots to factorize multi-granularity reconstruction states into various modality-aligned slots. These slots define different semantic directions that capture behavioral evidence across modalities and granularities (ranging from fine-grained events to high-level behavioral intents). Given a scenario query, a lightweight router first emphasizes the relevant slot semantic directions through an input-level residual pathway, filtering and fusing the behavioral information relevant to the query. This design enables scenario-specific evidence adaptation while preserving general-purpose user semantics, producing a compact scenario-adaptive embedding with a single dLLM forward pass. To provide grounded supervision for routing, we further construct \textbf{\bqa{}} (Behavioral Question-Answering), a large-scale corpus that pairs diverse scenario queries with behavior-grounded answers derived from traceable real-world behavioral logs.

Specifically, \ours{} is driven by two core mechanisms:
\begin{itemize}
\item \textbf{Meta-Slots for Denoising \& Factorization.} Learnable meta-slots organize diverse denoising states within the dLLM into transferable cross-modal semantics and modality-specific behavioral slots, effectively capturing the complex semantics of heterogeneous user logs.
\item \textbf{Residual Dynamic Router for Routing \& Compression.} A pre-fusion router conditions on the behavioral signal and scenario query to enhance relevant slots via a residual pathway. The dLLM then integrates this filtered information with the complete behavioral sequence and compresses it into a compact, scenario-adaptive representation.
\end{itemize}

Extensive experiments across ten industrial scenarios, alongside two large-scale online A/B tests, demonstrate that \ours{} achieves strong empirical performance in both unified and scenario-adaptive settings. Beyond these empirical gains, our analyses illustrate how massive and noisy behavioral sequences are associated with different downstream objectives via explicit query-slot alignments. This work effectively pushes the boundaries of dLLMs beyond their conventional role in language generation. By unlocking their iterative denoising process as a multi-granularity reconstruction space, we demonstrate their potential as powerful user representation learners.

\section{Related Work}
\paragraph{LLM for User Embedding.}
User embedding compresses heterogeneous user signals, including profiles, behavioral sequences, transaction records, and structured attributes, into compact representations reusable across downstream scenarios. Early methods capture sequential dependencies through recurrent or masked prediction objectives~\cite{ni2018perceive,sun2019bert4rec,fu2023robust} and improve transferability through contrastive learning across behaviors, contexts, and domains~\cite{lin2022dual,zou2022multi,dou2025transferable}. Recent LLM-based methods serialize heterogeneous signals as behavioral tokens and learn unified representations through next-token or next-behavior prediction~\cite{zhu2024liber,gao2025instruction,li2026panther}. Several studies further replace causal attention with bidirectional masking to better integrate long and heterogeneous behavior sequences~\cite{yuan2026decoder,li2025conan}. Query-adaptive methods condition representations on natural-language task descriptions, allowing the same profile to support different downstream objectives~\cite{yuan2026query}. However, they typically apply query conditioning or alignment only after heterogeneous behaviors have been fused and compressed by the backbone. Query adaptation can therefore modify the final embedding but has limited ability to recover modality-specific evidence underutilized or discarded during earlier representation formation.

\paragraph{Diffusion Language Models.}
Diffusion language models (dLLMs) progressively corrupt and recover discrete tokens through iterative denoising~\cite{austin2021structured,sahoo2024simple}. Their bidirectional architecture incorporates global context, achieving strong performance in language generation and reasoning~\cite{nie2026large,ye2025dream,you2026llada}. Some work shows that intermediate states encode useful timestep-dependent information for specific domains like math~\cite{wang2025time,liu2026mixture}. Moreover, recent embedding and retrieval studies typically pool only the final bidirectional states into fixed representations~\cite{zhang2025diffusion,eslami2026diffusion,wang2026diffretriever}. However, these text-centric advances leave the potential of dLLMs for heterogeneous user representation largely unexplored. We introduce a \emph{reconstruction-space} perspective for diffusion-based user modeling: rather than treating denoising merely as token recovery, \ours{} leverages behavioral and semantic targets to produce complementary multi-granularity states, factorizes them into latent slots, routes query-relevant evidence, and compresses it into a compact user embedding.
\section{Preliminaries}
\label{sec:preliminaries}
\paragraph{User Representation Learning (URL).} 
\label{sec:problem_statement}
Given a rolling user profile $\mathcal{X}=\left\{\mathbf{X}^{(m)}\right\}_{m=1}^{M}$, where $\mathbf{X}^{(m)}=\{x_{i}^{(m)}\}_{i=1}^{N^{(m)}}$ contains behavioral tokens from modality $m$, URL employs a model $f$ to learn a user representation $\mathbf{z}$. In a large payment platform, these modalities span diverse sources like billing records, navigation paths, search histories, app usage, campaign interactions, and structured attributes~\cite{ni2018perceive,hou2022towards,yang2023empowering}. 
Furthermore, $f$ must serve diverse downstream scenarios, each defined by a natural-language query $\mathbf{Q}$ specifying its objective. While the underlying profile $\mathcal{X}$ remains fixed, different scenarios depend on substantially different evidence. Therefore, in realistic industrial systems, the conventional approach of compressing the user profile into a single holistic representation $\mathbf{z}$ is often inadequate for adapting to diverse scenarios.
To this end, a multi-scenario user model $f_{\psi}$ must preserve diverse evidence relevant to different queries. Formally, given $\mathcal{X}$ and $\mathbf{Q}$, our objective is to learn a \emph{query-aware user representation} as $\mathbf{z}(\mathbf{Q})
=
f_{\psi}\!\left(\mathcal{X},\mathbf{Q}\right)
\in\mathbb{R}^{d_z}$,
where $\mathbf{z}(\mathbf{Q})$ remains compact and reusable while adapting the emphasized evidence to the scenario described by $\mathbf{Q}$. This creates a fundamental tension between a user profile infrastructure and scenario-dependent information requirements.

\paragraph{Masked Diffusion Language Modeling.}
\label{sec:masked_diffusion}
Let $\mathbf{x}_0=(x_{0,1},\ldots,x_{0,L})$ denote a clean token sequence at timestep 0.
Masked diffusion~\cite{austin2021structured,sahoo2024simple,nie2026large,ye2025dream} defines a forward noising process that gradually replaces tokens with a special \texttt{[MASK]} token. At timestep $t$, the sequence $\mathbf{x}_t\sim q(\mathbf{x}_t\mid\mathbf{x}_0)$ is obtained using a timestep-dependent masking rate. Let $\Omega_t$ denote the masked positions in $\mathbf{x}_t$.
A diffusion language model $F_{\theta}$ receives $\mathbf{x}_t$ and timestep $t$, and predicts the original tokens at the masked positions $\Omega_t$. It is trained using
\begin{equation}
\mathcal{L}_{\mathrm{MDM}}
=
-\mathbb{E}_{\mathbf{x}_0,t,\mathbf{x}_t}
\left[
\frac{1}{|\Omega_t|}
\sum\nolimits_{i\in\Omega_t}
\log p_{\theta}\!\left(x_{0,i}\mid\mathbf{x}_t,t\right)
\right].
\label{eq:masked_diffusion_objective}
\end{equation}
Unlike autoregressive prediction, masked diffusion allows each position to interact with visible context in both directions. Its contextualized hidden states, $
\mathbf{H}_t=F_{\theta}(\mathbf{x}_t,t),
\label{eq:diffusion_states}
$
also vary with the noise level and masking pattern. 

\section{Methodology}
\label{sec:method}

A static and holistic user representation makes it difficult to filter the relevant information required for different scenarios, where sparse but decisive evidence (e.g., an unusual transaction relevant to fraud detection) may consequently be diluted by large amounts of unrelated browsing or app activity.
Although masked diffusion models~\cite{nie2026large} provide rich bidirectional states, directly pooling their final hidden states still collapses these signals into a static and entangled representation. The key challenge is therefore to preserve heterogeneous behavioral signals and selectively expose them when a concrete scenario query arrives. We address this challenge with \ours{}, a diffusion state routing framework that separates evidence preservation from scenario-dependent evidence selection. 

\subsection{A Diffusion State Routing Framework}
\label{sec:diffusion_state_routing}

We first illustrate the overall architecture of \ours{} in Figure~\ref{fig:arch}, which follows a two-stage learning paradigm.
In \textbf{Stage I (\emph{Denoise} \& \emph{Factorize})}, we partially mask behavioral tokens and fully mask answer tokens. Learnable meta-slots then factorize these diverse reconstruction views into transferable cross-modal and modality-specific slots. In \textbf{Stage II (\emph{Route} \& \emph{Compress})}, the input combines clean behavioral and query embeddings with the learned slots. Guided by the query, a lightweight router emphasizes relevant slot directions via an input-level residual pathway. Finally, a single dLLM forward pass integrates these filtered signals with the full sequence to output a compact, scenario-adaptive representation. This design enables scenario-specific adaptation while preserving general-purpose user semantics.

To train our framework and explicitly supervise query-dependent routing, we construct \textbf{\bqa{}}\footnote{The construction of high-quality and diverse question-answer pairs is detailed in the appendix.}, a large-scale behavioral QA corpus. Unlike conventional labels that only provide prediction outcomes, \bqa{} connects scenario objectives directly to supporting behaviors. Each triplet $\left(\mathcal{X}, \mathbf{Q}, \mathbf{A}\right)$ includes an $L_a$-length answer sequence $\mathbf{A}=(a_{1},\ldots,a_{L_a})$ summarizing scenario-relevant evidence, yielding traceable semantic targets for Stage I query-grounded reconstruction and Stage II representation alignment.

\begin{figure*}
    \centering
    \includegraphics[width=1\linewidth]{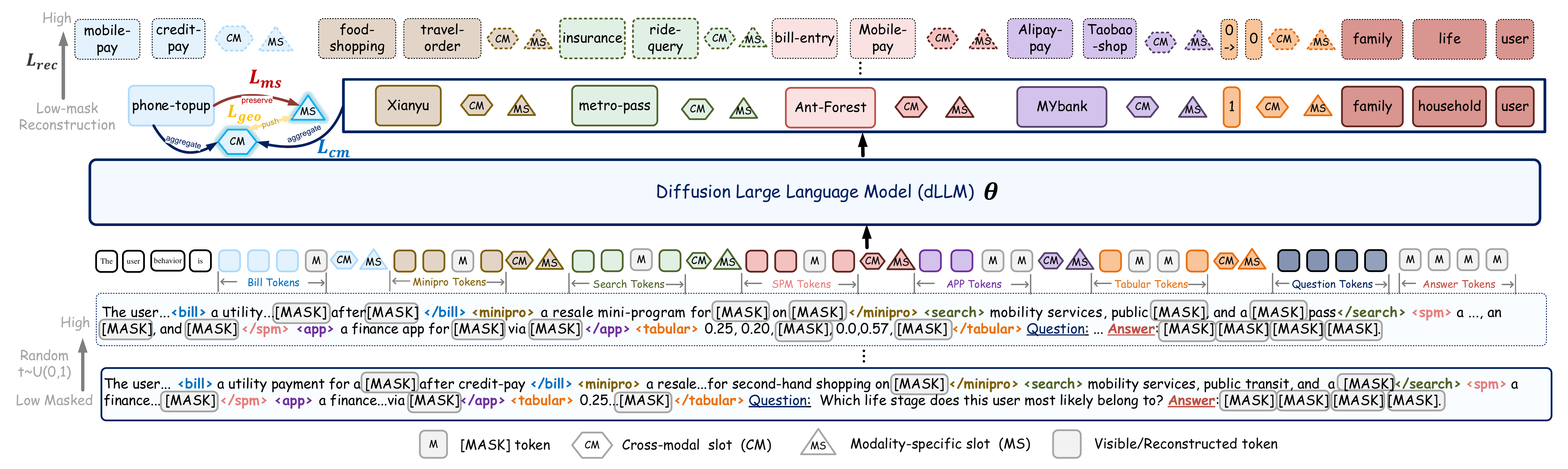}
    \caption{ Illustration of Stage I in \ours{}.
For each instance, \ours{} samples a timestep $t \sim \mathcal{U}(0,1)$, partially masking behavioral tokens and fully masking the answer, while keeping the query and meta-slots visible. The bidirectional dLLM jointly reconstructs both targets. Using the Bill modality in a low-mask pass ($t \approx 0$) as an example, we show how contextualized modality-specific (MS) slots preserve modality-level denoising state, whereas cross-modal (CM) slots aggregate various semantics. Repeating this process across random timesteps progressively organizes diverse reconstruction states into these slot factors.
}
    \label{fig:overall}
\end{figure*}

\subsection{Stage I: Denoise and Factorize}
\label{sec:denoise_factorize}

\paragraph{Denoising.}
Individual behavioral tokens $\mathbf{X}^{(m)}$ describe historical activities, and a behavior-grounded answer $\mathbf{A}$ summarizes the complete profile from the perspective of a particular scenario query $\mathbf{Q}$. 
For each modality $m$, we deploy $K$ learnable slots as $\mathbf{S}^{(m)}  
=
\left[
\mathbf{s}_{0}^{(m)}\!=\!\mathbf{s}_{\mathrm{cm}}^{(m)};  
\mathbf{s}_{1}^{(m)};  
\ldots;
\mathbf{s}_{K-1}^{(m)}  
\right]$,
where $\mathbf{s}_{0}^{(m)}\!=\!\mathbf{s}_{\mathrm{cm}}^{(m)}$ is a cross-modal semantic slot across various source and $\{\mathbf{s}_{k}^{(m)}\}_{k=1}^{K-1}$ retain modality-specific patterns within each source. These meta-slots define distinct semantic directions within the multi-granularity semantic space.
Then, we alternately attach the $K$ learnable meta-slots after their modality-specific behavioral tokens $\mathbf{X}^{(m)}$, followed by the query and answer, as the input to the user encoder $F_\theta$ to model the token-to-slot mapping: $\mathbf{X}^{\mathrm{in}} =
\left[
\mathbf{X}^{(1)};
\mathbf{S}^{(1)};  
\ldots;
\mathbf{X}^{(M)};
\mathbf{S}^{(M)};  
\mathbf{Q};
\mathbf{A}
\right]$.
At each training step, we sample a timestep $t \sim \mathcal{U}(0,1)$ and partially mask exactly $\max(1, \lfloor \rho_m(t) |\mathbf{X}^{(m)}| \rfloor)$ behavioral tokens of modality $m$ using a linear schedule $\rho_m(t)\in[0, 1]$. Across iterations, this stochastic sampling covers the time-indexed reconstruction conditions. All answer tokens are fully masked, while the query and meta-slots remain visible. Fully masking the answer prevents semantic leakage and forces the model to recover it from the profile according to the visible query.


Let $\Omega_t^{(m)}$ denote the masked positions in modality $m$, and let $\widetilde{\mathbf{X}}_{t}^{\mathrm{in}}$ denote the noisy version of the complete Stage-I input sequence $\mathbf{X}^{\mathrm{in}}$ obtained at diffusion noise level $t$. The model jointly reconstructs the missing behaviors and the complete answer:
\begin{align}
\mathcal{L}_{\mathrm{rec}}
=
-\mathbb{E}_{(\mathcal{X},\mathbf{Q},\mathbf{A}),t}\; 
\Big[
\frac{1}{L_a}
\sum\nolimits_{j=1}^{L_a}
\log p_{\theta}\!\left(
 a_{j}\; 
 \mid
 \widetilde{\mathbf{X}}_{t}^{\mathrm{in}},t\; 
\right) + \\
\frac{1}{M}
\sum\nolimits_{m=1}^{M}
\frac{1}{|\Omega_t^{(m)}|}
\sum\nolimits_{i\in\Omega_t^{(m)}}
\log p_{\theta}\!\left(
 x_{i}^{(m)}\; 
 \mid
 \widetilde{\mathbf{X}}_{t}^{\mathrm{in}},t\; 
\right)
\nonumber
\Big].
\label{eq:asymmetric_reconstruction}
\end{align}
The two reconstruction targets provide complementary supervision. Across sampled noise levels, light corruption retains detailed local evidence, whereas heavier corruption requires broader cross-token and cross-source inference from the remaining context. In parallel, answer reconstruction integrates the partially observed profile according to the scenario query and provides an explicit global semantic target. Denoising thus exposes complementary behavioral reconstruction states while separately grounding query-relevant global semantics through the answer.

\paragraph{Multi-granularity meta-slot factorization.}
We use learnable meta-slots as semantic anchors to organize these states into a small set of reusable factors.
For modality $m$, the learnable slot set $\mathbf{S}^{(m)}$ is jointly contextualized with the behavioral profile by the dLLM.
At the sampled noise level $t$, we collect the contextualized hidden states at the corresponding slot positions as $\mathbf{H}_{t}^{(m)}$
We introduce two complementary objectives to organize these slots into modality-specific behavioral factors and cross-modal semantic factors learned from the diverse reconstruction space.

First, we encourage the modality-specific slots to capture diverse denoising states within their corresponding modalities.
Let $\mathbf{H}_{t}^{\mathbf{X};(m)}$ denote the behavioral denoising states of modality $m$.
We use the contextualized modality-specific slot states $\mathbf{H}_{t,[1:K-1]}^{\mathbf{S};(m)}$ to reconstruct the corresponding behavioral denoising states: $\widehat{\mathbf{H}}_{t}^{\mathbf{X};(m)}
=
\operatorname{Attn}\!\left(
\mathbf{H}_{t}^{\mathbf{X};(m)},
\mathbf{H}_{t,[1:K-1]}^{\mathbf{S};(m)},
\mathbf{H}_{t,[1:K-1]}^{\mathbf{S};(m)}
\right)$.
The corresponding objective is
\begin{equation}
\mathcal{L}_{\mathrm{ms}}
=
\frac{1}{M}
\mathbb{E}_{(\mathcal{X},\mathbf{Q},\mathbf{A}),t}
\left[
\sum\nolimits_{m=1}^{M}
\left\|
\operatorname{sg}\!\left(\mathbf{H}_{t}^{\mathbf{X};(m)}\right)
-
\widehat{\mathbf{H}}_{t}^{\mathbf{X};(m)}
\right\|_{F}^{2}
\right].
\label{eq:local_loss}
\end{equation}
Here, $\operatorname{sg}(\cdot)$ denotes the stop-gradient operation.
This objective encourages the modality-specific slots to preserve the information required to recover the behavioral states of each modality.

The second objective uses the fully masked answer to supervise the cross-modal slot states. We pool the reconstructed answer hidden states as
$
\mathbf{g}_{t}
=
\operatorname{Pool}\!\left(
\mathbf{H}_{t}^{\mathbf{A}}
\right),
\label{eq:answer_semantics}
$
and aggregate the contextualized cross-modal slot states across modalities as $\overline{\mathbf{h}}_{t,0}^{\mathbf{S}} = \frac{1}{M}\sum\nolimits_{m=1}^{M}\mathbf{H}_{t,[0]}^{\mathbf{S};(m)}$.
Their alignment is optimized by
\begin{equation}
\mathcal{L}_{\mathrm{cm}}
=
\mathbb{E}_{(\mathcal{X},\mathbf{Q},\mathbf{A}),t}
\left[
1-
\cos\!\left(
\overline{\mathbf{h}}_{t,0}^{\mathbf{S}},
\operatorname{sg}\!\left(\mathbf{g}_{t}\right)
\right)
\right].
\label{eq:global_loss}
\end{equation}
Because the answer is reconstructed by integrating observable evidence across modalities according to the scenario query, this objective encourages the cross-modal slots to capture global, query-relevant semantics from multiple behavioral modalities.

\begin{figure}
    \centering
    \includegraphics[width=1\linewidth]{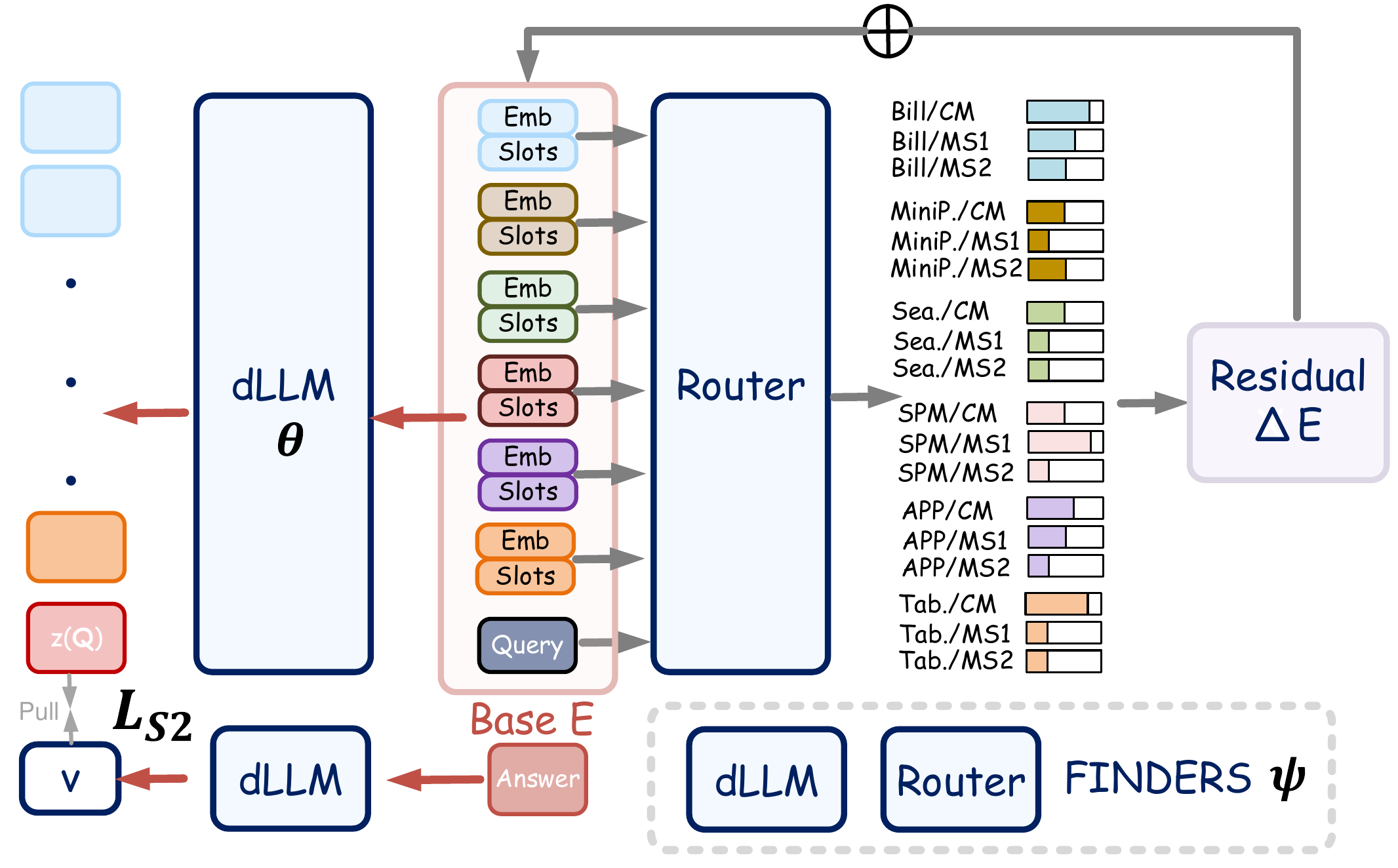}
\caption{
\textbf{Illustration of Stage II in \ours{}.}
Given the clean input, the router computes query-conditioned weights over modality-slot pairs to form a residual $\Delta\mathbf{E}$. A single dLLM forward pass then processes this enhanced sequence into the user representation $\mathbf{z}$, which is contrastively aligned with the answer $\mathbf{v}$ during training.
}
    \label{fig:router}
        \vspace{-0.2cm}
\end{figure}

\paragraph{Factor diversity.}

Without an explicit constraint, different slots may repeatedly encode the same dominant behavior, leaving few distinct factors for downstream routing. We row-normalize the contextualized slot states $\mathbf{H}_{t}^{\mathbf{S};(m)}$ into $\widetilde{\mathbf{H}}_{t}^{\mathbf{S};(m)}$ and impose the following diversity constraint:
\begin{equation}
\mathcal{L}_{\mathrm{geo}}
=
\frac{1}{M}
\mathbb{E}_{(\mathcal{X},\mathbf{Q},\mathbf{A}),t}
\left[
\sum\nolimits_{m=1}^{M}
\left\|
\widetilde{\mathbf{H}}_{t}^{\mathbf{S};(m)}
\widetilde{\mathbf{H}}_{t}^{\mathbf{S};(m)\top}
-
\mathbf{I}_{K}
\right\|_F^2
\right].
\label{eq:slot_geometry}
\end{equation}
Applying this constraint to the contextualized slot states encourages different slots to specialize in distinct semantic regions.
The complete Stage I objective is 
\begin{equation}
\mathcal{L}_{\mathrm{S1}}
=
\mathcal{L}_{\mathrm{rec}}
+
\lambda_{\mathrm{ms}}\mathcal{L}_{\mathrm{ms}}
+
\lambda_{\mathrm{cm}}\mathcal{L}_{\mathrm{cm}}
+
\lambda_{\mathrm{geo}}\mathcal{L}_{\mathrm{geo}}.
\label{eq:stage1}
\end{equation}
We first optimize $\mathcal{L}_{\mathrm{rec}}$ to establish stable denoising states and then progressively increase the weights of the slot-specific objectives to organize these states into reusable factors.

\subsection{Stage II: Route and Compress}
\label{sec:route_compress}

After factorizing the reconstruction space into cross-modal and source-specific slots, Stage II uses the query to emphasize decision-relevant slots via routing.
\paragraph{Query-conditioned slot routing.}
We remove diffusion corruption and answer tokens while retaining the same modality-slot-query layout for clean representation learning. Each behavioral sequence $\mathbf{X}^{(m)}$ is mapped by its corresponding source-specific encoder into dense embeddings $\mathbf{E}^{(m)}$ in the dLLM input space, while the scenario query is mapped into query embeddings $\mathbf{E}^{\mathbf{Q}}$. These behavioral embeddings are alternately interleaved with the meta-slot sets $\mathbf{S}^{(m)}$, followed by the query embeddings, to form the clean base input: $\mathbf{E}^{\mathrm{base}}
=
\left[
\mathbf{E}^{(1)};
\mathbf{S}^{(1)};
\ldots;
\mathbf{E}^{(M)};
\mathbf{S}^{(M)};
\mathbf{E}^{\mathbf{Q}}
\right].$
The router pools the query embeddings into $\mathbf{e}^{\mathbf{Q}}$ and the behavioral embeddings of each modality into $\mathbf{e}^{\mathbf{X};(m)}$. Then, we compute a gate for each slot as $g^{(m)}_{k}(\mathbf{Q})
=
\sigma\!\left(
\operatorname{MLP}\!\left(
[\mathbf{e}^{\mathbf{X};(m)};
 \mathbf{s}_{k}^{(m)};
 \mathbf{e}^{\mathbf{Q}}]
\right)\Big/\tau\right)$.
The gate jointly considers the current behavior $\mathbf{e}^{\mathbf{X};(m)}$, the slot factor $\mathbf{s}_{k}^{(m)}$, and the scenario query $\mathbf{e}^{\mathbf{Q}}$. It therefore estimates the relevance of each cross-modal or modality-specific factor to the current decision objective. Since every slot is associated with a modality, these gates naturally capture both source relevance and evidence granularity.

\paragraph{Dual-mode residual adaptation.}
The router does not replace or prune the base profile but constructs a query-conditioned residual that strengthens scenario-relevant slot directions. The base pathway preserves the complete behavioral context and transferable user semantics, while the residual pathway adjusts how the dLLM reads this context for a particular scenario.
Then, the routed residual block is $\Delta\mathbf{E}^{(m)}(\mathbf{Q})
=
\left[
\mathbf{0}_{\mathbf{E}^{(m)}};
 g^{(m)}_{\mathrm{0}}(\mathbf{Q})\mathbf{s}_{\mathrm{0}}^{(m)};
 \ldots;
 g^{(m)}_{K-1}(\mathbf{Q})\mathbf{s}_{K-1}^{(m)}
\right],$
where $\mathbf{0}_{\mathbf{E}^{(m)}}$ leaves the behavioral embeddings unchanged. The gate at index $0$ controls the cross-modal semantic direction, while the remaining gates govern the modality-specific behavioral directions. Since all slots remain present in the base pathway, routing only adds query-dependent emphasis without discarding shared semantics or behavioral evidence.
As shown in Fig.~\ref{fig:router}, the residual blocks are concatenated as $\mathbf{E}^{\mathrm{route}}=
\left[
\Delta\mathbf{E}^{(1)}(\mathbf{Q});
\ldots;
\Delta\mathbf{E}^{(M)}(\mathbf{Q});
\mathbf{0}_{\mathbf{Q}}
\right]$ and added to the base input $\widetilde{\mathbf{E}}
=
\mathbf{E}^{\mathrm{base}}
+
\gamma\mathbf{E}^{\mathrm{route}}$.
The scaling coefficient $\gamma$ is learnable and initialized to zero, allowing Stage II to begin from the Stage-I base pathway and progressively introduce query-dependent adaptation during training.
Each slot is effectively scaled by $1+\gamma g^{(m)}_{k}(\mathbf{Q})$ before entering the dLLM, adjusting the emphasis of the learned factors before they interact with the current profile. The additive residual preserves the complete behavioral sequence, supports both general-purpose transfer and scenario-specific views, and avoids information loss from hard routing.

Finally, the residual-enhanced input is processed through one clean dLLM forward pass.
Let $\mathcal{I}^{\mathbf{Q}}$ denote the query-token positions. We compute $\mathbf{z}({\mathbf{Q}})
=
\operatorname{Norm}\!\left(
\mathbf{W}_{z}
\operatorname{Pool}\!\left(
\mathbf{H}[\mathcal{I}^{\mathbf{Q}}]
\right)
\right)$, where $\mathbf{H}
=
F_{\theta}\!\left(
\widetilde{\mathbf{E}},
t=0
\right)$.
Through bidirectional attention, the routed slots are contextualized by the current user's behaviors, while the query states aggregate the complete profile with greater emphasis on the selected factors. Different queries can therefore produce different representations for the same user without requiring separate encoders or representation stores.

\paragraph{Contrastive representation learning.}
Following common contrastive representation learning~\cite{yuan2026query,zhang2025diffusion}, the dLLM-based answer branch maps $\mathbf{A}_i$ into the normalized target
$
\mathbf{v}_i
=
\operatorname{Norm}\!\left(
F_{\theta}(\mathbf{A}_i)
\right).
$
The routed user embedding is treated as the anchor, its matched answer as the positive, and other in-batch answers as negatives:
\begin{equation}
\mathcal{L}_{\mathrm{S2}}
=
-\frac{1}{B}
\sum\nolimits_{i=1}^{B}
\log
\frac{
\exp\!\left(
\operatorname{sim}(\mathbf{z}(\mathbf{Q})_{i},\mathbf{v}_i)/\tau_c
\right)
}{
\displaystyle
\sum\nolimits_{j=1}^{B}
\exp\!\left(
\operatorname{sim}(\mathbf{z}(\mathbf{Q})_{i},\mathbf{v}_j)/\tau_c
\right)
}.
\label{eq:stage2}
\end{equation}
This objective teaches the routed representation to retain the behavioral evidence expressed by the matched answer.

\paragraph{Inference.}
At inference, \ours{} takes $\mathcal{X}$ and $\mathbf{Q}$ as input. It supports both scenario-adaptive (using $\mathbf{Q}$ with residual routing) and general-purpose (using a generic query with $\gamma=0$) serving. Both modes use a single forward pass through the shared encoder to yield vectors of identical dimensionality.

\begin{table*}[!t]
\centering
\scriptsize
\caption{AUC on ten industrial scenarios. Methods are grouped by their effective representation architecture. Prompt Tuning and Routing denote scenario-adaptive variants. Best results are in \textbf{bold} and second-best results are \underline{underlined}.}
\label{tab:main_results}
\resizebox{\textwidth}{!}{%
\begin{tabular}{lccccccccccc}
\toprule
& \multicolumn{4}{c}{\textbf{Marketing \& Payment Conversion}}
& \multicolumn{2}{c}{\textbf{Risk Control}}
& \multicolumn{2}{c}{\textbf{Financial Services}}
& \multicolumn{2}{c}{\textbf{Ecosystem Engagement}}
& \textbf{Avg.} \\
\cmidrule(lr){2-5}\cmidrule(lr){6-7}\cmidrule(lr){8-9}\cmidrule(lr){10-11}
\textbf{Method}
& \textbf{Promo} & \textbf{PayGift} & \textbf{NPay} & \textbf{Value}
& \textbf{Theft} & \textbf{Fraud}
& \textbf{Fund} & \textbf{Jiebei}
& \textbf{GamePay} & \textbf{Forest} & \textbf{AUC} \\
\midrule

\multicolumn{12}{@{}l@{}}{
\textit{\textbf{Bidirectional Encoder Architectures (BERT-style)}}} \\

Llama-Embed-Nemotron-8B
& 0.7040 & 0.6620 & 0.6410 & 0.8420
& 0.7520 & 0.9100 & 0.8110 & 0.8910
& 0.8750 & 0.9020 & 0.7990 \\

MSDP~\cite{fu2023robust}
& 0.7240 & 0.6740 & 0.6510 & 0.8760
& 0.7420 & 0.9280 & 0.8190 & 0.9070
& 0.8960 & 0.9180 & 0.8135 \\

One4all~\cite{shin2021one4all}
& 0.7280 & 0.6800 & 0.6570 & 0.8790
& 0.7460 & 0.9310 & 0.8220 & 0.9100
& 0.8990 & 0.9210 & 0.8173 \\

FOUND~\cite{dou2025transferable}
& 0.7440 & 0.6960 & 0.6740 & 0.8930
& 0.7570 & 0.9410 & 0.8370 & 0.9210
& 0.9100 & 0.9300 & 0.8303 \\

\midrule
\multicolumn{12}{@{}l@{}}{
\textit{\textbf{Causal/Autoregressive Architectures (AR)}}} \\

Qwen2.5-0.5B-Instruct~\cite{qwen2.5}
& 0.5880 & 0.5520 & 0.5380 & 0.7120
& 0.6620 & 0.8260 & 0.7010 & 0.8040
& 0.7710 & 0.8240 & 0.6978 \\

Qwen3-Embedding-0.6B~\cite{zhang2025qwen3}
& 0.6010 & 0.5660 & 0.5510 & 0.7260
& 0.6790 & 0.8390 & 0.7190 & 0.8180
& 0.7860 & 0.8370 & 0.7122 \\

KaLM-Embedding-Gemma3-12B~\cite{hu2025kalm} 
& 0.6490 & 0.6050 & 0.5880 & 0.7830
& 0.7180 & 0.8760 & 0.7660 & 0.8520
& 0.8280 & 0.8670 & 0.7532 \\

CPC~\cite{oord2018representation}
& 0.7310 & 0.6820 & 0.6600 & 0.8810
& 0.7490 & 0.9320 & 0.8250 & 0.9120
& 0.9020 & 0.9230 & 0.8197 \\

\textsc{Q-Anchor}~\cite{yuan2026query}
& 0.7831 & 0.7335 & 0.7049 & 0.9253
& 0.7830 & 0.9553 & 0.8615 & 0.9380
& 0.9326 & 0.9478 & 0.8565 \\

\textsc{Q-Anchor} + Prompt Tuning
& 0.7868 & 0.7374 & 0.7073 & 0.9270
& 0.7852 & 0.9571 & 0.8640 & 0.9402
& 0.9347 & 0.9500 & 0.8590 \\

\midrule
\multicolumn{12}{@{}l@{}}{
\textit{\textbf{Diffusion Language Models (dLLM)}}} \\

{Qwen2.5-Coder-0.5B-Instruct-diffusion-mdlm-v0.1}~\cite{zhou2026dllm}
& 0.5740 & 0.5410 & 0.5260 & 0.6980
& 0.6510 & 0.8150 & 0.6880 & 0.7920
& 0.7600 & 0.8120 & 0.6857 \\

Vanilla-dLLM
& 0.7876 & 0.7382 & 0.7078 & 0.9274
& 0.7859 & 0.9576 & 0.8646 & 0.9408
& 0.9351 & 0.9504 & 0.8595 \\

\ours{} ($\gamma=0$)
& \underline{0.7902} & \underline{0.7410}
& \underline{0.7089} & \underline{0.9288}
& \underline{0.7872} & \underline{0.9590}
& \underline{0.8660} & \underline{0.9420}
& \underline{0.9366} & \underline{0.9520}
& \underline{0.8612} \\

\ours{} + Routing
& \textbf{0.7938} & \textbf{0.7452}
& \textbf{0.7106} & \textbf{0.9311}
& \textbf{0.7887} & \textbf{0.9610}
& \textbf{0.8672} & \textbf{0.9437}
& \textbf{0.9383} & \textbf{0.9540}
& \textbf{0.8634} \\
\bottomrule
\end{tabular}}
\end{table*}

%

\begin{figure}
    \centering
    \includegraphics[width=1\linewidth]{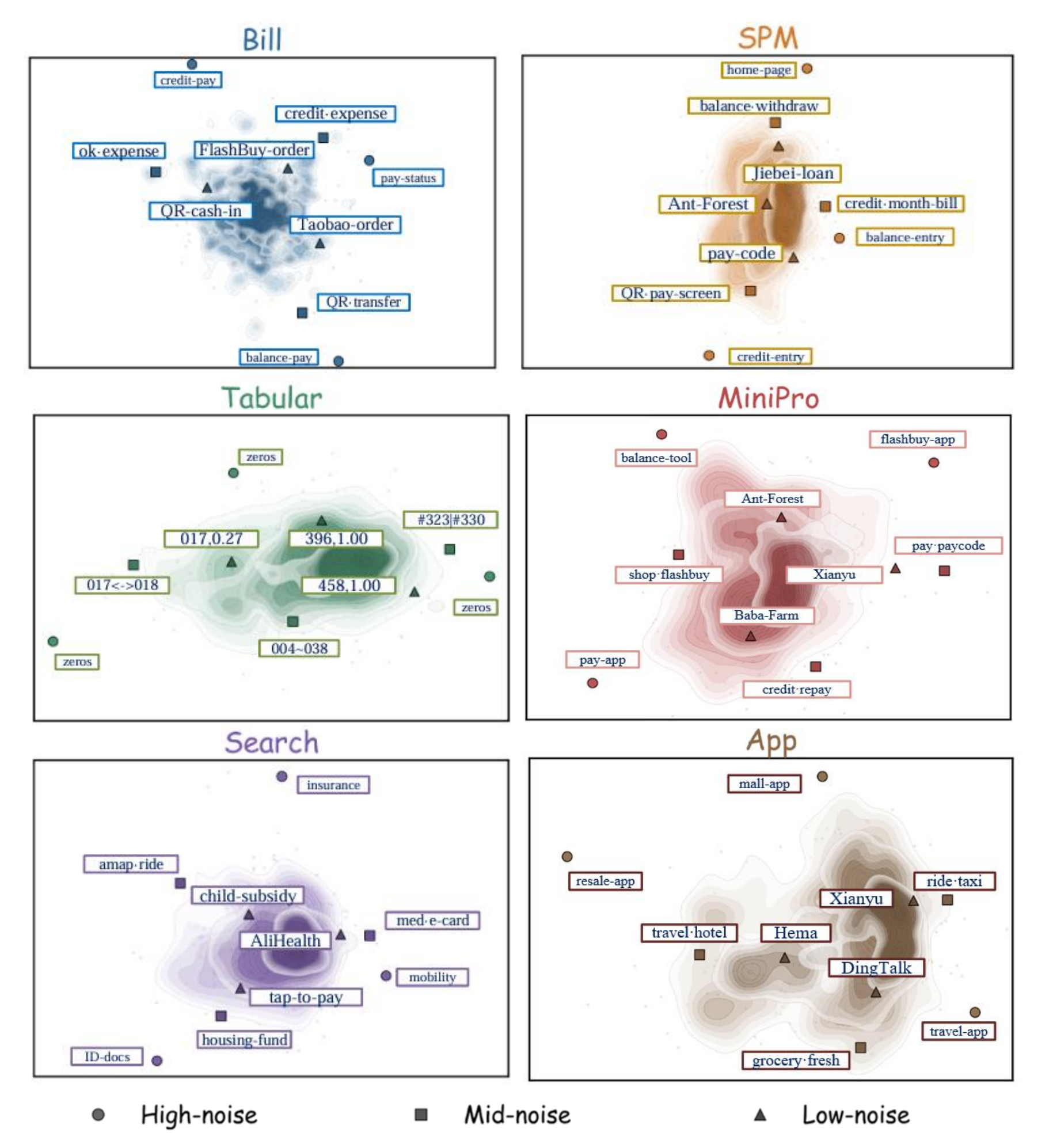}
    \caption{Visualization of Stage-I diffusion-state semantics.
Semantic distributions and high-frequency tokens are shown
across modalities under low-, middle-, and high-noise states. Different marker shapes denote the three noise regimes.}
    \label{fig:word}
        \vspace{-0.5cm}
\end{figure}

\section{Experiments}
\label{sec:experiments}

\paragraph{Data and Task Scenarios.}
All user-trained methods are pretrained on $\sim$$4.096\times10^9$ anonymized rolling profiles from a large payment platform. Each profile contains six asynchronous behavioral sources: \textbf{(1)} billing and transactions (Bill), \textbf{(2)} Mini Program activities (MiniPro), \textbf{(3)} explicit searches (Search), \textbf{(4)} page exposures and navigation (SPM), \textbf{(5)} app-level browsing and purchases (App), and \textbf{(6)} structured attributes (Tabular). Pretraining and evaluation sets are separated by both user and time. 
We evaluate ten binary classification tasks ($\sim$$5\times10^5$ test users each) across four domains: \textbf{(1) Marketing and Payment Conversion} (Promo, PayGift, NPay, Value), relying on complex exposure-intent-payment chains; \textbf{(2) Risk Control} (Theft, Fraud), requiring sparse anomaly detection amid high-volume routine activity; \textbf{(3) Financial Services} (Fund, Jiebei), depending on stable asset, credit, and cash-flow evidence; and \textbf{(4) Ecosystem Engagement} (GamePay, Forest), involving multi-step activities across apps, pages, and Mini Programs.
\paragraph{Baselines.}
We group the baselines by their attention architectures: 
\textbf{(1) Bidirectional encoders (BERT-style)} allow tokens to attend to both preceding and subsequent evidence. This includes Llama-Embed-Nemotron-8B~\cite{babakhin2025llama}, MSDP~\cite{fu2023robust}, One4all~\cite{shin2021one4all}, and FOUND~\cite{dou2025transferable} (featuring bidirectional multimodal fusion). 
\textbf{(2) AR models} impose a strict left-to-right constraint. We include off-the-shelf Qwen2.5-0.5B-Instruct~\cite{qwen2.5}, Qwen3-Embedding-0.6B~\cite{zhang2025qwen3}, KaLM-Embedding-Gemma3-12B~\cite{hu2025kalm}, CPC~\cite{oord2018representation}, and \textsc{Q-Anchor}~\cite{yuan2026query}. \textsc{Q-Anchor} trains Qwen2.5-0.5B on industrial behaviors via AR and contrastive objectives, with a variant learning scenario-specific soft prompts. 
\textbf{(3) dLLMs } overcome AR constraints via bidirectional masked denoising. We include the off-the-shelf Qwen2.5-Coder-0.5B-Instruct-diffusion-mdlm-v0.1~\cite{zhou2026dllm} and Vanilla-dLLM, which follows our setup and pools final query-token states. Finally, $\ours{}$ ($\gamma=0$) performs meta-slot factorization without routing, and $\ours{}$+Routing adds query-conditioned residual routing for scenario adaptation.

\paragraph{Implementation and Evaluation Protocols.}
Following~\cite{yuan2026query}, source-specific \texttt{gte-base} encoders~\cite{li2023towards} map behavioral sources into dense tokens. \ours{} adopts Qwen2.5-Coder-0.5B-Instruct-diffusion-mdlm-v0.1~\cite{zhou2026dllm} as its backbone. All user-trained methods share the pretraining split, token budget, optimization schedule, and 128-dimensional output. Models are trained for 200k steps on 128 A100-80G GPUs using AdamW (batch size 2,048, peak LR $2\times10^{-4}$, cosine decay). We apply LoRA ($rank=64, \alpha=64$) and assign $K=8$ meta-slots per source. 
In \emph{unified mode} ($\gamma=0$), a generic query generates a shared, frozen user vector for all tasks. In \emph{scenario-adaptive mode} (\ours{}+Routing), task-specific queries activate the pretrained router. For both evaluation settings, we freeze all representation components and train only a linear prediction head per task. We report AUC~\cite{bradley1997use}. More details are in the appendix.

\subsection{Main Performance}
\label{sec:main_results}

As shown in Table~\ref{tab:main_results}, both \ours{} ($\gamma=0$) and \ours{}+Routing outperform substantially larger general-purpose models despite using a 0.5B backbone. This highlights that for sparse, structured industrial behaviors, domain-aligned pretraining is more critical than sheer model scale. Notably, the user-pretrained Vanilla-dLLM surpasses the AR baseline (\textsc{Q-Anchor}), confirming that bidirectional reconstruction better models asynchronous, partially ordered logs than strict causal serialization. 
Moreover, state factorization and routing effectively address the limitation of premature compression. In the base pathway ($\gamma=0$), meta-slots organize the multi-granularity reconstruction space into transferable cross-modal and modality-specific factors, preserving diverse evidence that Vanilla-dLLM otherwise collapses into an entangled representation. By routing these learned slots via an input-level residual pathway, \ours{} leverages explicit rationales from the \bqa{} corpus to selectively emphasize scenario-relevant semantics before cross-source fusion. 
This yields a highly effective scenario-adaptive representation, particularly beneficial for complex conversion tasks like Promo and PayGift. Because predictive signals here span long exposure-intent-payment chains, \ours{}+Routing achieves absolute AUC improvements of +0.70\% and +0.78\% over the best scenario-adaptive AR baseline (\textsc{Q-Anchor} + Prompt Tuning). Broadly, \ours{} bridges the gap between a universal user representation and scenario-dependent evidence requirements, allowing a single centrally served representation to dynamically adapt to diverse queries without separate encoders or representation stores.
\begin{figure}
    \centering
    \includegraphics[width=1\linewidth]{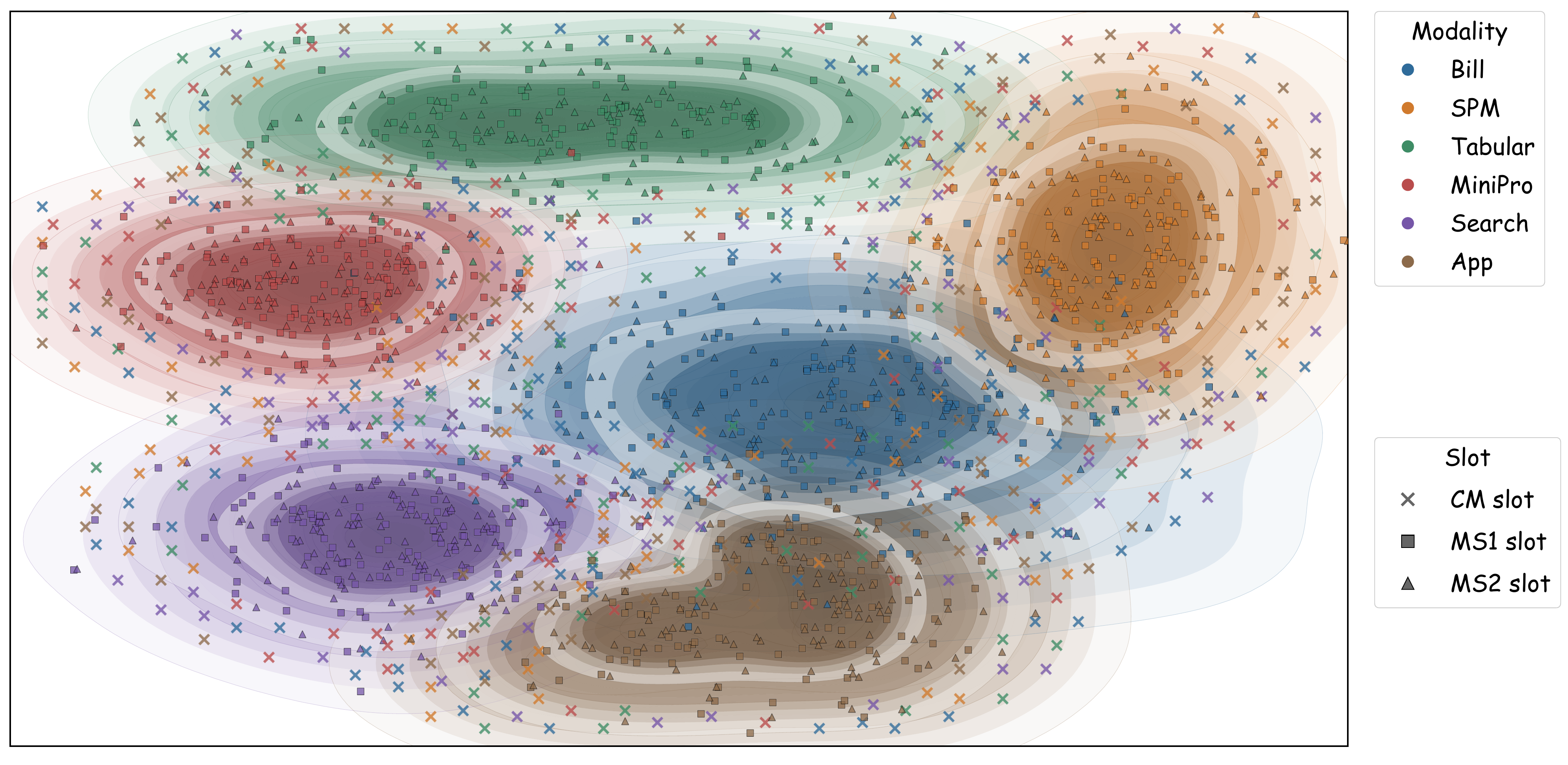}
    \caption{  t-SNE visualization of low-, middle-, and high-noise
state distributions together with meta-slot states extracted
from Stage-I forward passes. Colors denote modalities, while different marker
shapes indicate the cross-modal slot (CM) and two modality-specific slots
(MS1 and MS2).}
    \label{fig:slot_specialization}
    \vspace{-0.1cm}
\end{figure}

\subsection{Analysis}

\begin{tcolorbox}[boxsep=0mm, left=1.2mm, right=1.2mm, colframe=black!55, colback=black!5]
\textbf{Takeaway 1:} The diffusion reconstruction space preserves complementary evidence at multiple granularities, reconciling fine-grained behavior modeling with holistic user understanding.
\end{tcolorbox}

\begin{figure}[t]
    \centering
    \includegraphics[width=0.9\linewidth]{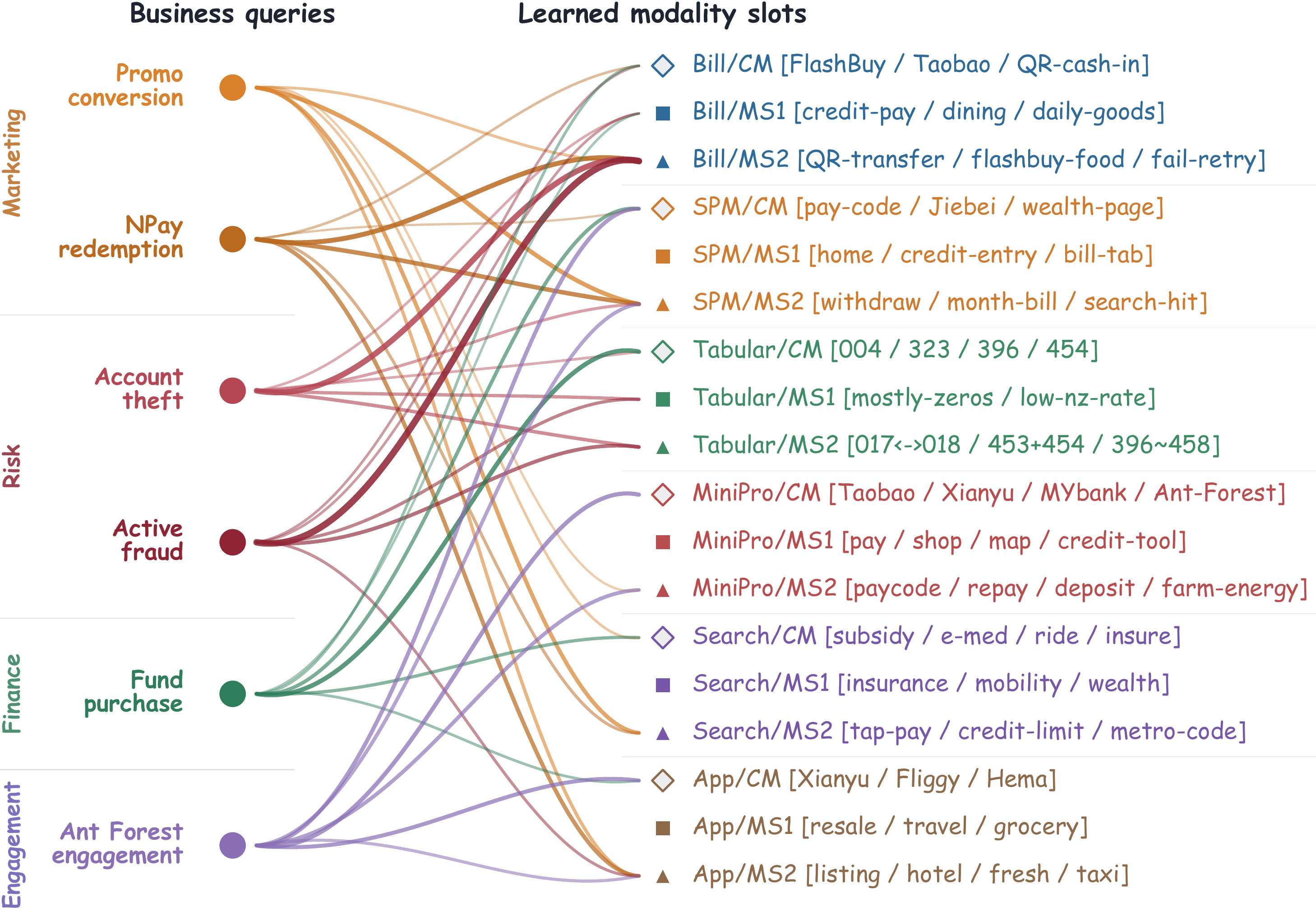}
    \caption{Query-conditioned routing over modality-indexed slot directions. Left-node colors denote business domains, right-node colors denote the behavioral modality associated with each slot, and marker shapes indicate slot types (CM, MS1, and MS2). Routing weights are aggregated over randomly sampled test instances.}
    \label{fig:query_slot_routing}
    \vspace{-0.5cm}
\end{figure}

\textbf{\emph{Multi-Granularity Evidence Space.}} 
To examine the learned reconstruction space, we freeze the Stage-I model and extract hidden states of each behavioral source at low, middle, and high noise levels. As shown in Figure~\ref{fig:word}, the high-frequency tokens associated with these states suggest a structured semantic evolution rather than redundant views. Specifically, low-noise states (circles) anchor to concrete atomic events (e.g., individual orders and app interactions), whereas middle-noise states (squares) capture compositional patterns (e.g., navigation-to-payment chains). Under high noise (triangles), representations become broader, reflecting coarse semantic prototypes like travel demands or general payment preferences. 
These patterns reflect a tendency rather than a predefined timestep-to-granularity assignment. As denoising proceeds, ambiguous semantic regions progressively resolve into precise modality-specific evidence. Importantly, the three regimes overlap without collapsing, indicating that atomic events, cross-event dependencies, and abstract semantics coexist at different resolutions. This evolution is also modality-dependent: \textit{Bill} retains fine-grained transaction semantics, while \textit{SPM} and \textit{Search} expose intent and navigation transitions. These observations validate our \emph{denoise} design: the reconstruction space exposes complementary behavioral views, allowing subsequent factorization and routing to selectively access scenario-relevant information.

\begin{figure*}[t]
    \centering
    \includegraphics[width=0.9\linewidth]{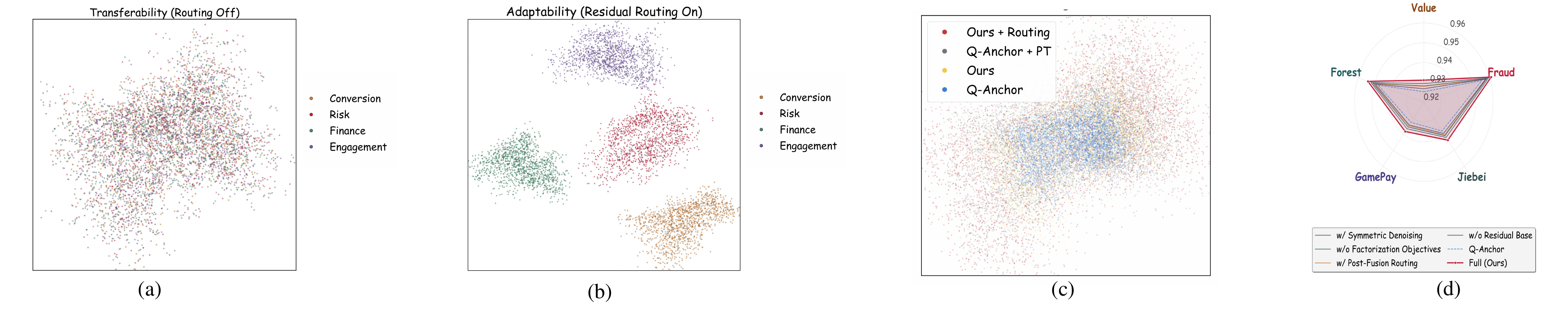}
    \caption{Transferability and scenario adaptation. t-SNE visualization of the embeddings (a) with routing disabled and (b) with routing enabled. (c) Representation-space comparison across methods. (d) Task-wise AUC results of the ablation study.}
    \label{fig:dual_mode_validation}
\end{figure*}

\textbf{\emph{Structured Evidence Factorization.}}
We examine whether meta-slots organize the multi-granularity reconstruction space rather than acting as redundant pooling tokens. As shown in Figure~\ref{fig:slot_specialization}, modality states retain distinct manifolds, while contextualized slots exhibit complementary specialization. Cross-modal slots (crosses) lie near shared boundaries, consistent with capturing globally supported semantics. Modality-specific slots (squares for MS1, triangles for MS2) remain close to their source manifolds and occupy distinct subregions, reflecting local, compositional behavioral patterns. This geometry validates our \emph{factorization} design: global alignment promotes cross-source semantics, local reconstruction preserves source-specific evidence, and geometric regularization prevents slot collapse. So these slots organize the diffusion-state space into structured evidence directions reusable for subsequent routing.

\begin{tcolorbox}[boxsep=0mm, left=1.2mm, right=1.2mm, colframe=black!55, colback=black!5]
\textbf{Takeaway 2:} By weighting modality-indexed slot directions according to business queries, the router reveals which evidence sources and granularities support different decision objectives.
\end{tcolorbox}

\textbf{\emph{Interpretability.}} To understand how the router selectively exposes these learned factors across different semantic granularities, we visualize the query-conditioned routing distributions. As shown in Figure~\ref{fig:query_slot_routing}, different scenarios assign high routing weights to distinct modality-indexed slot directions. For example, Promo conversion queries capture short-term intent by emphasizing factors such as \textit{SPM/MS2} (withdraw, month-bill, search-hit). But Fund purchase queries prioritize global semantic factors such as \textit{Tabular/CM} and Ant Forest engagement captured by \textit{MiniPro/MS2} ( farm-energy).
 Since each slot is associated with a behavioral source, these routing weights jointly reflect source preference and evidence granularity. The representative behaviors shown beside each channel describe the semantic region anchored by that slot, providing an interpretable link between business queries and supporting behavioral evidence. Such evidence-aware routing offers a transparent view of how massive and noisy user behaviors support different downstream intents.
\begin{tcolorbox}[
    boxsep=0mm,
    left=1.2mm,
    right=1.2mm,
    colframe=black!55,
    colback=black!5
]
\textbf{Takeaway 3:}
Dual-pathway compression reconciles transferability and scenario adaptation. A shared base pathway preserves general user semantics, while an input-level residual pathway exposes query-dependent scenario views.

\end{tcolorbox}
\textbf{\emph{Reconciling Shared Representation and Diverse Scenarios.}}
We examine whether \ours{} resolves the tension of \emph{premature compression}: supporting diverse scenario-specific decisions from a single pre-computed profile. With routing disabled ($\gamma=0$), embeddings from different domains cluster in a shared manifold (Figure~\ref{fig:dual_mode_validation}a), demonstrating that the base pathway preserves broadly reusable semantics. Activating the routed residual induces query-dependent displacements over this space (Figure~\ref{fig:dual_mode_validation}b), adapting evidence emphasis without replacing the underlying representation. Consequently, the fully routed model spans a richer, more expressive scenario-conditioned space than its generic base and AR alternatives (Figure~\ref{fig:dual_mode_validation}c). Ablations (Figure~\ref{fig:dual_mode_validation}d) further confirm our design: removing the base pathway  weakens transferability, while relying on post-fusion routing or disabling slot factorization limits adaptation. Thus, the residual router successfully decouples stable user semantics from scenario-specific evidence reorganization, allowing a single encoder and representation store to dynamically serve heterogeneous objectives.

Overall, \ours{} advances user representation learning beyond static, entangled compression. Diffusion states expose complementary behavioral views across multi-granularity noise levels, meta-slots organize these views into reusable semantic directions, and residual routing emphasizes scenario-relevant evidence prior to cross-source fusion. This demonstrates the potential of dLLMs for adaptive user modeling by exploiting reconstruction states to meet diverse downstream requirements.

\subsection{Industrial Online A/B Testing}
\label{sec:online_evaluation}
We evaluate \ours{} through two 14-day online A/B tests on a large-scale payment platform. Traffic allocation, downstream models, and decision policies remain strictly identical between the treatment and the control. Unless otherwise specified, reported metrics denote relative improvements over the baseline.
\textit{\textbf{1) Scenario I: Digital Payment Marketing Conversion.}} 
This scenario focuses on identifying users likely to complete a payment following exposure to digital marketing campaigns. Compared with the production baseline, \ours{} achieves an \textbf{absolute AUC improvement of 1.18 percentage points} (from 0.9443 to 0.9561) and a \textbf{62.5\% relative reduction in CTCVR calibration error} (PCOC deviation from 1.0 decreases from 0.0301 to 0.0113). These dual improvements indicate that our method effectively captures complex conversion signals, enhancing both ranking and calibration.
\textit{\textbf{2) Scenario II: Active Fraud Identification.}}
This risk scenario requires sparse anomalies to remain salient amid high-volume routine activities. \ours{} improves the \textbf{KS score by 1.34\%} and \textbf{recall at a fixed FPR by 1.92\%}, while reducing the \textbf{risk loss rate by 0.71\%} and \textbf{manual audit load by 3.15\%}. This demonstrates that our method prevents the dilution of decisive fine-grained evidence, boosting both detection quality and operational efficiency.
\textit{\textbf{3) Deployment at Production Scale.}}
The deployed service incrementally refreshes user profiles, returning a standard 128-dimensional embedding seamlessly compatible with existing downstream models. Crucially, inference requires no answer tokens, noise injection, or iterative denoising, and residual routing produces the scenario-adaptive embedding via \emph{a single clean dLLM forward pass}. Overall, these online gains prove that preserving heterogeneous evidence before compression and dynamically reorganizing it contextually yields production value.

\section{Conclusion}

We present \ours{}, a router-equipped dLLM framework designed for industrial user representation. By preserving and selectively routing multi-granularity diffusion states, \ours{} empowers a shared encoder to support both general-purpose transferability and query-dependent scenario adaptation. Across various scenarios and online tests, \ours{} achieves the best performance among the evaluated baselines in both unified and adaptive user modeling. The diffusion reconstruction space serves as a qualitatively inspectable interface, organizing heterogeneous behavioral evidence across different semantic granularities and explicitly connecting massive behavioral logs with downstream decision intents. This perspective elevates user representation learning from the static pooling of implicit signals to a dynamic, evidence-aware process, providing a robust foundation for building adaptive representation infrastructures in real-world intelligent systems. Broadly, our findings highlight the immense potential of diffusion language models as good, general-purpose representation learners, extending their utility beyond purely generative tasks.
\newpage
\clearpage
\bibliographystyle{ACM-Reference-Format}
\bibliography{custom}

\end{document}